\documentclass[11pt,a4paper]{article}

\newif\iflinenos
\linenosfalse   

\usepackage{lineno}

\PassOptionsToPackage{table}{xcolor}

\usepackage[utf8]{inputenc}
\usepackage[T1]{fontenc}
\usepackage{lmodern}

\usepackage[top=2.5cm,bottom=2.5cm,left=2.5cm,right=2.5cm]{geometry}
\usepackage{setspace}
\usepackage[numbers,super,sort&compress,comma]{natbib}

\usepackage[hidelinks,breaklinks=true]{hyperref}
\usepackage{url}

\usepackage{graphicx}
\graphicspath{{figures/}}

\usepackage{booktabs}
\usepackage{multirow}
\usepackage{xcolor}
\usepackage{siunitx}
\usepackage{array}
\usepackage{pdflscape}   

\usepackage{amsmath}

\usepackage[font=small,labelfont=bf,margin=0.5cm]{caption}
\usepackage{subcaption}

\usepackage{microtype}
\usepackage{parskip}
\usepackage{enumitem}

\colorlet{rowColorGood}{green!15}
\colorlet{rowColorBest}{green!45}

\usepackage{titlesec}
\titleformat{\section}{\large\bfseries}{\thesection.}{0.5em}{}
\titleformat{\subsection}{\normalsize\bfseries}{\thesubsection.}{0.5em}{}

\begin{document}

\iflinenos
	\linenumbers
\fi

\makeatletter
\AddToHook{env/figure/begin}{\iflinenos\internallinenumbers\fi}
\AddToHook{env/table/begin}{\iflinenos\internallinenumbers\fi}
\makeatother

\begin{center}
{\LARGE\bfseries Climate-based Pre-screening of Self-sustaining Regreening Opportunities in Drylands: A Case Study for Saudi Arabia}

\vspace{1.2em}

{\normalsize
Katja~Froehlich\textsuperscript{$\dagger$,1},\quad
Jonathan~Klein\textsuperscript{$\dagger$,1},\quad
Ibrahim~S.~Elbasyoni\textsuperscript{2,3},\quad
Julian~D.~Hunt\textsuperscript{2},\quad
Yoshihide~Wada\textsuperscript{2},\quad
Dominik~L.~Michels\textsuperscript{1}
}

\vspace{0.6em}

{\small
\textsuperscript{1}Computational Sciences Group, KAUST, Thuwal, KSA\\[2pt]
\textsuperscript{2}Plant Science Program, KAUST, Thuwal, KSA\\[2pt]
\textsuperscript{3}Crop Science Department, Faculty of Agriculture, Damanhour University, Damanhour, Egypt
}

\vspace{0.4em}

{\small
\textsuperscript{$\dagger$}Co-first author
}

\vspace{0.4em}
\end{center}

\vspace{0.8em}
\hrule
\vspace{0.8em}

\begin{center}
{\bfseries Abstract}
\end{center}

\noindent
Large-scale restoration in drylands is widely promoted to address land degradation and biodiversity loss, yet many efforts rely on long-term irrigation, limiting sustainability in water-scarce regions. A key challenge is identifying locations where native vegetation can persist without intensive management while minimizing costly field campaigns. A scalable pre-screening framework is presented that integrates climate and remote sensing data to enable cost-efficient site selection in arid environments using Saudi Arabia as a case study. A Climate Suitability Score (CSS), derived from machine learning models trained on expert-curated reference sites, captures complex climatic dependencies on vegetation persistence. Using multi-year ERA5-Land data for Saudi Arabia, national-scale prediction maps are generated and combined with vegetation indices to identify areas where climate is favorable, but vegetation remains underdeveloped. Multi-criteria screening reduces candidates to thirteen priority locations. Climatically analogous intact ecosystems provide benchmarks for restoration targets and indicate that an average 2.5 fold increase in vegetation coverage is a realistic target for restoration efforts. Overall, this approach narrows the search space, reduces costs, and supports resilient ecosystem recovery planning in water-limited regions.

\vspace{0.8em}
\hrule
\vspace{1em}

\section{Introduction}

Desertification is defined as land degradation in hyper-arid to dry sub-humid areas caused mainly by land mismanagement or adverse climatic conditions, e.g., overgrazing and drought\cite{UNCCD:1999,VicenteSerrano:2024}. These stressors lead to loss of vegetation, which destabilizes the soil layers and exposes the land to erosion\cite{SaezSandino:2024}. Desertification leads to long-term destruction of native ecosystems and, subsequently, to the displacement of populations that rely on these ecosystems\cite{MartinezValderrama:2020}. As a result, human intervention is crucial for the restoration and maintenance of native ecosystems\cite{Hohl:2020}. Land restoration efforts, specifically regreening, have spurred global projects to combat desertification, including the Bonn Challenge, the Great Green Wall launched by the African Union, the Three Norths Shelterbelt Program in China, and the Saudi Green Initiative\cite{Deng:2024,Parr:2024}. However, restoration success in water-limited environments remains highly variable, and many projects fail to establish persistent vegetation cover or require continued human intervention to remain viable. A recurring limitation is that restoration planning often prioritizes visible short-term greening rather than long-term ecological persistence. These projects often rely on continued intensive maintenance, such as irrigation with extracted groundwater\cite{Cernansky:2021}.

As restoration commitments expand across drylands, there is an urgent need to distinguish locations capable of supporting self-sustaining vegetation recovery from those that remain dependent on external inputs\cite{Hohl:2020,Parr:2024}. Otherwise, precious resources are wasted, and the existing ecosystem could even degrade further\cite{Parr:2024,Turner:2023}. In the past, many studies have been conducted to find efficient solutions for regreening; however, they have mostly focused on specific, relatively small regions\cite{Takoutsing:2023,Eshetie:2025,PiriSahragard:2021}. For large-scale restoration projects, not all regions are equally suitable for regreening, necessitating a complex selection process. One option is to select locations based on \emph{Geographic Information System} (GIS) maps, using metrics such as the \emph{Normalized Difference Vegetation Index} (NDVI) to assess changes in vegetation cover at large scales\cite{AlAmri:2026,Hasan:2024}. A reduction of the NDVI over time indicates a loss of vegetation that could potentially be restored through regreening initiatives.

Generally, NDVI values between 0.1 and 0.2 are associated with barren land, while values between 0.2 and 1 correspond to vegetation\cite{Khalil:2024}. However, in arid climates, NDVI analysis can be misleading because vegetation naturally has low visible cover due to smaller leaf areas despite extensive root systems, making standard thresholds unreliable\cite{Khalil:2024,AlmalkiR:2022,Kirschner:2021}. Additionally, NDVI cannot effectively differentiate between orchards and natural perennial vegetation, often leading to overestimation of natural vegetation potential. Finally, low NDVI values are ambiguous, as they may represent either true desert conditions or degraded land that could still be suitable for land restoration.

Rather than relying solely on the NDVI, suitable locations for land restoration should therefore be selected based on additional environmental factors such as climate, soil, and water quality, as well as accessibility for plant deployment and monitoring\cite{Hohl:2020,Parr:2024}. However, including these factors increases the cost of the selection analysis as it necessitates on-site analysis for accurate soil and water quality assessment.

In this work, we therefore propose to employ an additional pre-selection step that relies solely on climate factors, for which global datasets are readily available. For example, the ERA5 Land dataset provides globally consistent, multi-year climate data with high temporal resolution\cite{MunozSabater:2021}.

Since plant survival depends on a complex interplay of factors rather than on simple criteria such as specific temperature intervals, it is challenging to explicitly compute the long-term survival potential of a given location. Therefore, we opt for an implicit, machine learning-based approach, in which the training data captures the underlying complexity of the survival criteria. A model is trained on expert-curated reference locations representing persistent natural vegetation, degraded landscapes and climatically unsuitable areas. It then predicts a \emph{Climate Suitability Score for Vegetation Survival} (CSS). A promising land restoration location is then characterized by a high CSS but low current vegetation cover.

We apply our framework to Saudi Arabia as a model for dryland systems. Here, annual declines of 1.2--1.5\% in shrub and tree cover within rangelands have been reported\cite{AlAmri:2026,MinistryEnv:2023}. The Saudi Green Initiative pledges to restore 74 million ha of land by planting 10 billion trees across the country as part of its Vision 2030\cite{SGI:2024}. We generate national-scale opportunity maps, prioritize feasible restoration sites, and identify climatically analogous intact ecosystems that provide realistic reference conditions and restoration targets.

Our results establish a scalable approach for directing restoration effort toward landscapes with higher long-term ecological potential in water-limited regions worldwide.

\section{Results}

\subsection{Climate Suitability Score (CSS) Model Training}

\begin{figure}[tbp]
\centering
\includegraphics[width=0.82\textwidth]{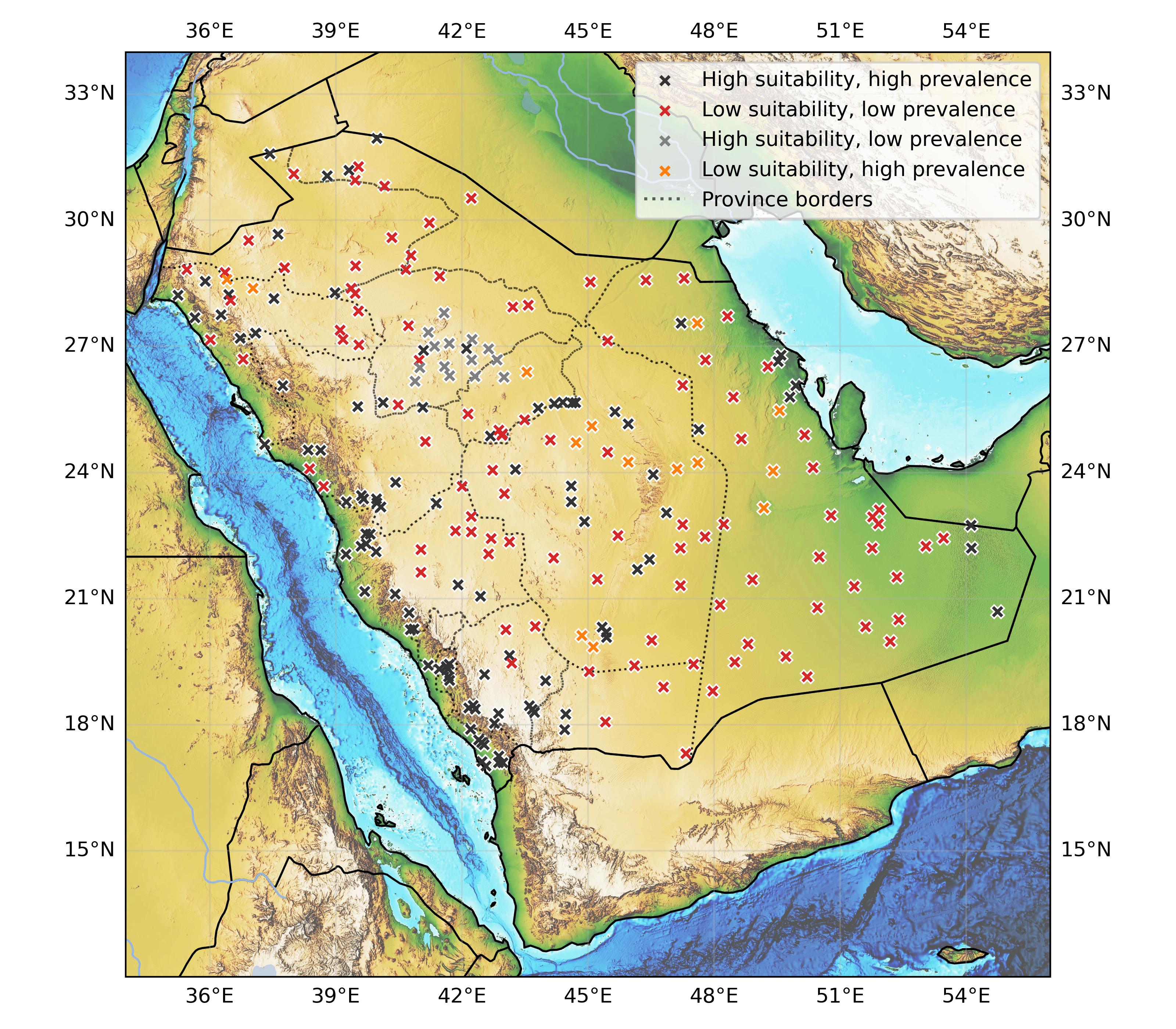}
\caption{\textbf{Map of Saudi Arabia.} The 230 sample locations and their classes are shown according to climatic suitability for vegetation and prevalence of vegetation.}
\label{fig:sample_positions}
\end{figure}

Saudi Arabia encompasses strong climatic gradients ranging from hyper-arid deserts to comparatively mesic mountain systems in the south-west, providing a stringent test case for climate-based restoration screening. We trained two complementary machine-learning models on expert-curated reference locations representing combinations of climatic suitability and vegetation prevalence. Multi-year ERA5-Land time series were transformed into compact feature representations and used to predict a continuous climate suitability score across the national landscape.

Saudi Arabia exhibits pronounced hydroclimatic heterogeneity, reflecting its predominantly arid to semi-arid climate and complex topography. Climatic and topographic constraints contribute to the patchy distribution of natural vegetation. Most vegetation cover is concentrated in the southwestern highlands, whereas in other regions it is largely restricted to wadis, stream networks, and other favorable microhabitats (Figure~\ref{fig:saudi_ndvi}). Extreme aridity, high solar radiation, and sandy soils create inhospitable conditions across much of the country, selecting species with high tolerance to heat and drought. To capture this variability, we constructed a nationwide training dataset comprising 230 locations distributed across all major climate zones and landforms (Figure~\ref{fig:sample_positions}, Figure~\ref{fig:saudi_ndvi}).

23 climate variables from the \emph{ERA5-Land} hourly data set\cite{MunozSabater:2021} were selected, and three-hourly data spanning the years 2020--2024 were extracted for each site. Only variables irrelevant for our region, such as \emph{snow density} or \emph{lake ice temperature}, were not present in our selection. This ensures that all potentially useful information is available, while less important variables are automatically assigned appropriate weights during the training process.

Since no explicit method exists for assigning CSS values to climate vectors, we initially used binary values (1\,=\,high, 0\,=\,low) to label the training data. The locations were selected according to four different categories: A high potential was assigned to locations that either display five years of continuous vegetation coverage (according to the NDVI value) or to locations that have been described as deteriorated by anthropogenic activities in the literature\cite{AlRowaily:2018,Salih:2021}. A low potential was assigned to locations that do not display any vegetation for five years, as well as to agricultural areas with no vegetation apart from the irrigated fields and plantations (Table~\ref{tab:training_categories}). Each sample location was additionally examined using high-resolution remote sensing imagery to address the limitations of NDVI analysis in arid climates and to ensure accurate assignment. The NDVI-based categories imply a correlation between CSS and current vegetation, which would be valid only in the absence of human interference. This correlation is broken by the inclusion of the other categories which include agricultural areas and degraded land.

\begin{table}[tbp]
\centering
\caption{\textbf{Sample category sizes and input labels for model training.} 230 training samples were chosen according to four categories. Binary labels were assigned according to the category.}
\label{tab:training_categories}
\begin{tabular}{llccc}
\toprule
 & & \multicolumn{2}{c}{Vegetation Prevalence} & Binary Label \\
\cmidrule(lr){3-4}
 & & \textbf{High} & \textbf{Low} & \\
\midrule
\multirow{2}{*}{Climatic Suitability} & \textbf{high} & 101 & 14 & \textbf{1} \\
 & \textbf{low} & 14 & 101 & \textbf{0} \\
\bottomrule
\end{tabular}
\end{table}

The CSS regression task is characterized by having a relatively low number of input samples (230 across all categories combined, Table~\ref{tab:training_categories}) but a large number of dimensions per sample. This makes selecting the appropriate machine learning model challenging. Here, the suitability of two different machine learning models was evaluated.

The \emph{Best Linear Unbiased Predictor} (BLUP) model has been used in quantitative genetics\cite{Habier:2007,OgetEbrad:2024}. There, a similar problem exists in estimating the breeding value of individual specimens using several million DNA markers. Second, a neural network (NN) classifier was used as a baseline comparison. Neural networks are well known as powerful tools for identifying complex patterns, however, training the networks robustly with only a small amount of available training data can be a major challenge\cite{Dsouza:2020}.

Before training the different models, the input vectors were compressed via Fourier coefficient selection for the BLUP model and via an autoencoder's latent vectors for the neural network. This enabled variable-sized inputs and thus a control over how much of the climatic information is utilized.

\begin{figure}[tbp]
\centering
  \includegraphics[width=\textwidth]{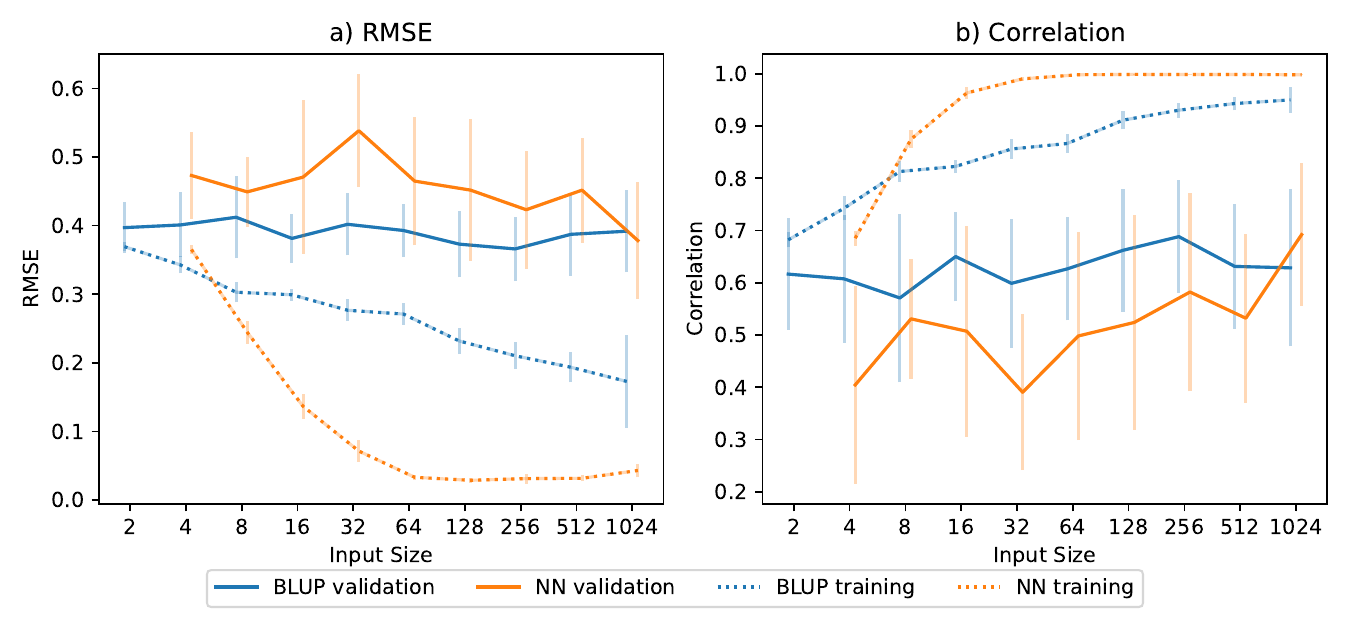}
\caption{\textbf{Model training results.} Training results for BLUP and neural network for different input sizes. Average results and error bars are shown for the 10 training repetitions of each size. The input size corresponds to ``number of Fourier coefficients per variable'' for BLUP and ``autoencoder latent vector size'' for neural network.}
\label{fig:training_results}
\end{figure}

The training was repeated for each input size 10 times, with a randomized 10\% of the dataset held back for performance validation each time. The \emph{Root Mean Square Error} (RMSE) is the natural evaluation metric for the neural network, as it was trained with an RMSE loss function, while the correlation value is a typical evaluation metric in BLUP-based analysis in genetics\cite{Habier:2007} (Figure~\ref{fig:training_results}).

Increasing the input size can increase performance on the training data but has a negligible effect on the performance on the validation data. Generally, performance converges at a size of 32 for BLUP and at a size of 256 for the neural network, indicating that increasing the input data size further would not translate into increased performance. Further, the neural network fits the training data better, whereas BLUP generalizes better (as evidenced by the slightly superior results on the validation set).

Overall, the evaluation shows no clear favorite---neither between BLUP and neural network, nor between the different input sizes. We therefore opted to use all trained models by independently evaluating them and averaging their results. This is known as ensemble learning and has been shown to be effective at avoiding overfitting and increasing robustness\cite{Black:2021,Azad:2025}.

\begin{figure}[tbp]
\centering
\includegraphics[width=\textwidth]{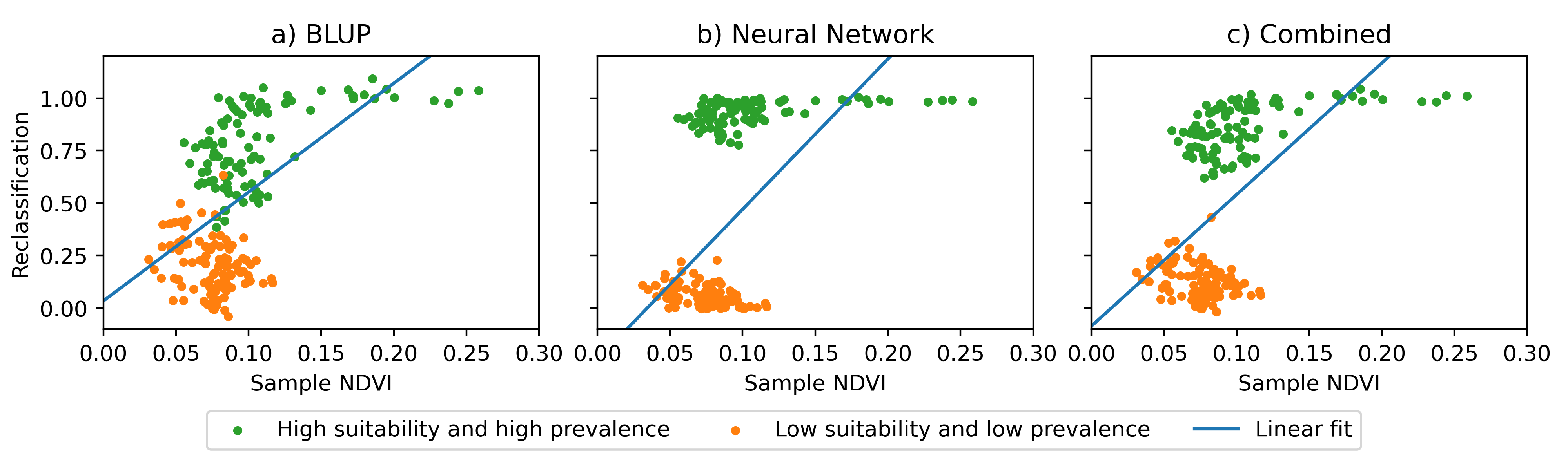}
\caption{\textbf{Correlation between sample reclassification of two categories and NDVI.} 101 high suitability / high vegetation prevalence (green) and 101 low suitability / low vegetation prevalence (orange). Training samples were reclassified after training with (a)~BLUP, (b)~neural network, and (c)~combined results.}
\label{fig:reclassification}
\end{figure}

While the input labels are binary, the classification results are continuous. This allows for a reclassification, where the input locations are assigned new labels based on their classification results from the different models (Figure~\ref{fig:reclassification}).

The BLUP reclassifications have a much wider spread compared to the neural network reclassifications, which show a much sharper separation between the high- and low-suitability categories. Additionally, the high-suitability samples have a much wider spread in their NDVI values across all models compared to the low-suitability samples (Figure~\ref{fig:reclassification}).

To find out if these high-suitability samples are classified consistently across different models, the samples were ranked by their individual scores for BLUP, the neural network, and their combination. We then analyzed the overlap between the 20 top- and bottom-ranked locations according to each score to estimate their reclassification agreement. Additionally, we included an NDVI-based ranking to estimate the correspondence between samples with a high predicted suitability and their NDVI values. The overlaps between these 4 sets of 20 locations are plotted in Figure~\ref{fig:overlap}.

These rankings show a significant agreement between the 3 classification methods (BLUP, neural network, and their combination). Almost the same 20 locations were ranked either top or bottom across all models. Furthermore, the results show a strong correspondence between high NDVI values and high suitability classification. In contrast, the sample locations with the lowest NDVI values only have a minimal overlap with the lowest-ranked locations (Figure~\ref{fig:overlap}).

This suggests that the CSS, independent from the machine learning model, is suitable to robustly identify climates that yield high NDVI values.

\subsection{Climate-based Suitability Modelling Identifies Persistent Vegetation Signals}

To predict the regions in Saudi Arabia with the highest likelihood of climate suitability for vegetation survival, the climate dataset for the whole country was evaluated using both trained BLUP and neural network models of all sizes.

\begin{figure}[tbp]
\centering
\includegraphics[width=\textwidth]{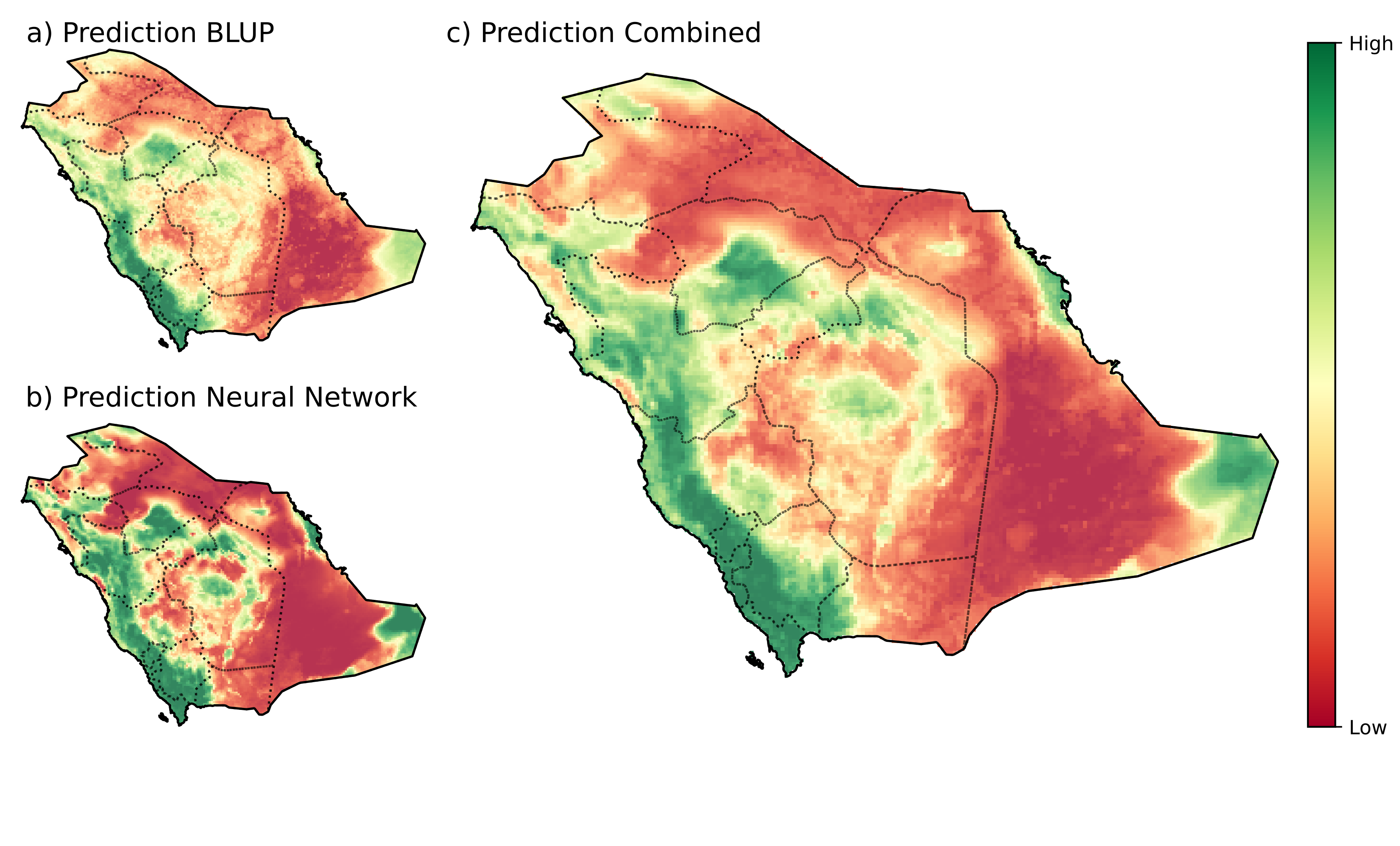}
\caption{\textbf{Climate Suitability Score (CSS) prediction maps.} (a, b) Combined maps over all input sizes for BLUP and neural network. (c) Combination of all BLUP and neural network maps.}
\label{fig:css_maps}
\end{figure}

Spatial prediction patterns from the aggregated BLUP models identified the highest likelihood of climate suitability in the southwest, extending from Jazan northward through Al Baha and into the southern Makkah region, spanning the Red Sea coastal plain and the Asir Mountains (Figure~\ref{fig:css_maps}a). The neural network models predicted a largely congruent core region, but with sharper class boundaries and additional areas of moderate-to-high CSS in Hail, central Riyadh, the eastern coastal zone, the southeast border with the UAE and Oman, Tabuk, and parts of northern Saudi Arabia (Figure~\ref{fig:css_maps}b). Combined model outputs yielded a more conservative classification (Figure~\ref{fig:css_maps}c), analogous to the ensemble effect observed during training reclassification (Figure~\ref{fig:reclassification}c). Despite their different model structures, the two approaches recovered broadly consistent suitability patterns and after thresholding the outputs of each model, they reach a relatively high agreement of 71\% (measured by the \emph{intersection over union} score, Figure~\ref{fig:iou}). With no clear favorite, we therefore used unweighted averaging to combine both. High scores were concentrated in regions known to support persistent natural vegetation, particularly the south-western highlands\cite{Ghazal:2015} and portions of the Red Sea escarpment\cite{Alharthi:2023}, whereas the most arid interior and south-eastern desert regions were consistently assigned low suitability. Agreement between independently trained models indicates that robust climatic signals underpin the predicted patterns rather than dependence on a single algorithmic approach.

\subsection{National-scale Opportunity Mapping Reveals Under-realized Restoration Zones}

\begin{figure}[tbp]
\centering
\includegraphics[width=\textwidth]{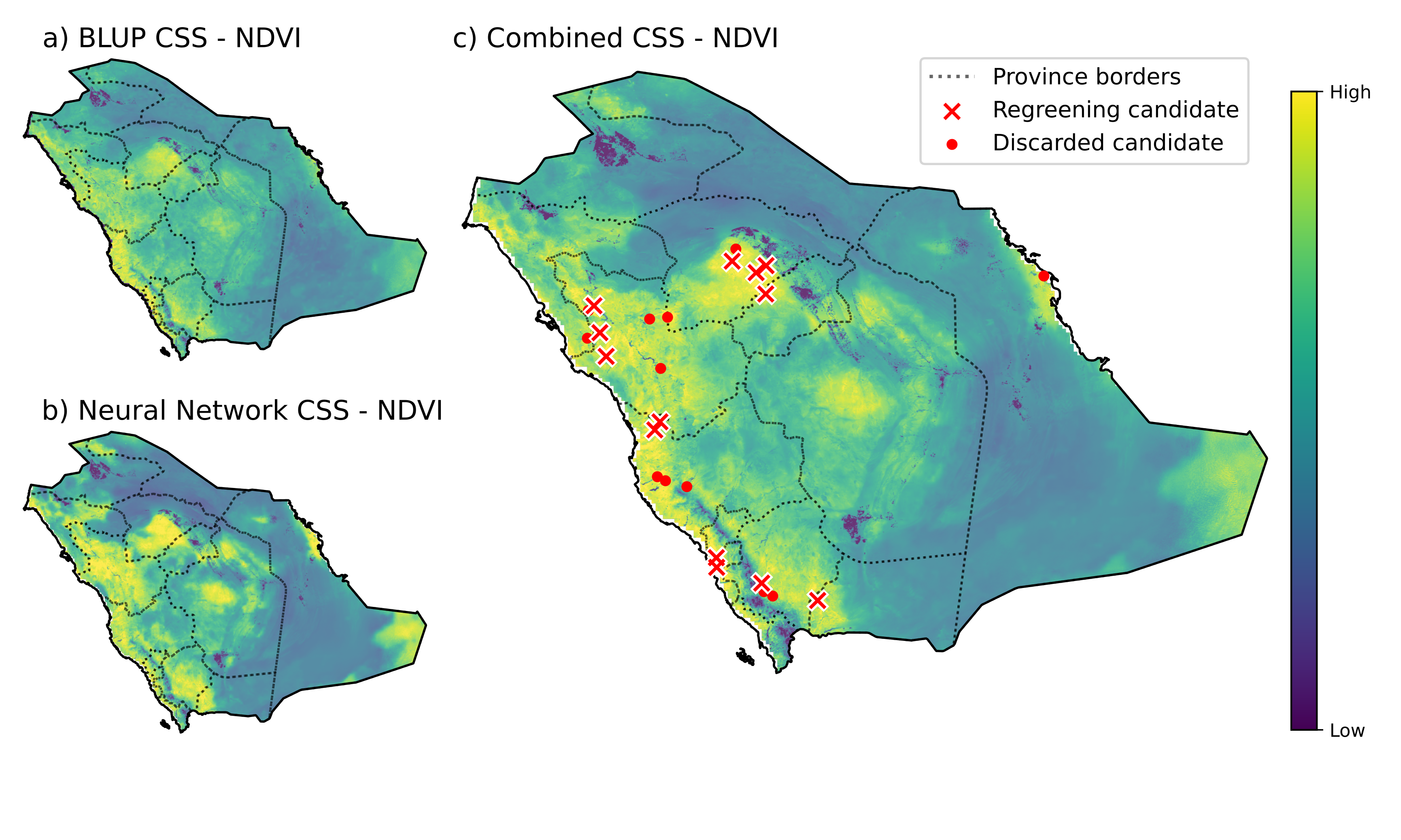}
\caption{\textbf{Comparison between CSS prediction and NDVI.} Difference maps between the CSS (Figure~\ref{fig:css_maps}) and NDVI (Figure~\ref{fig:saudi_ndvi}) for (a)~BLUP, (b)~neural network, and (c)~both combined. The 25 candidate locations for land restoration were selected based on the combined difference map (red marks).}
\label{fig:opportunity_map}
\end{figure}

Climatic suitability alone does not identify restoration priorities because highly suitable areas may already support substantial vegetation cover. Therefore, the predicted CSS was integrated with contemporary multi-year NDVI to detect landscapes where climate appears favorable but vegetation remains below expected levels. These ``opportunity zones'' represent locations where ecological recovery may be possible if non-climatic barriers such as historical degradation, grazing pressure or local disturbance are addressed. To identify candidate regions for land restoration, we intersected high climatic suitability predictions with low five-year average summer NDVI values, thereby excluding transient seasonal vegetation responses to precipitation. The resulting CSS--NDVI contrast map (Figure~\ref{fig:opportunity_map}) distinguishes areas with high NDVI but low climatic potential, corresponding predominantly to irrigated agricultural systems, and areas with high climatic suitability but low NDVI, indicating potential priority zones for natural vegetation establishment without continuous human aid.

This analysis highlighted candidate regions beyond the already vegetated south-west, including sections of the western corridor, inland Hail, parts of the central plateau and selected northern landscapes. In contrast, intensively cultivated zones with high vegetation cover but low climatic suitability were deprioritized, consistent with dependence on irrigation rather than natural persistence (Figure~\ref{fig:opportunity_map}). The combined framework therefore distinguishes restoration opportunity from both existing natural vegetation and agriculturally maintained greening.

\subsection{Multi-criteria Filtering Sharply Reduces the Search Space for Field Campaigns}

To convert broad opportunity maps into actionable targets, the 25 highest-ranked candidate locations (Figure~\ref{fig:opportunity_map}c) were evaluated based on elevation, terrain, climate zone, proximity to settlements, accessibility and visible vegetation structure from high-resolution imagery (Table~\ref{tab:candidate_locations}).

Vegetation diversity and patterns are influenced by proximity to wadis (temporary streams) and elevation\cite{Ghazal:2015,AlHarbi:2024}. Exposure to human activities, such as agriculture or the expansion of urbanization, is a competing factor for natural vegetation\cite{AlAmri:2026,Moatamed:2021}.

Based on topographical features, precipitation patterns, and temperature regimes, Saudi Arabia can be divided into several principal climatic regions: northern, Red Sea coastal, interior, Asir Highlands, East, and West Rub al-Khali\cite{Almazroui:2015} (Figure~\ref{fig:saudi_ndvi}). Each of these regions is subject to distinct atmospheric and geographic influences, resulting in pronounced heterogeneity in environmental conditions.

The prevailing climate across most of the country is arid, with precipitation largely confined to the winter months (November--April) and typically originating from warm waters of the Arabian Gulf\cite{Almazroui:2020}. Rainfall is highly variable and often sporadic. The Asir Highlands in the southwest form a notable exception, receiving higher precipitation due to orographic lifting and the influx of moist air, which sustains greater vegetation density. Annual mean temperatures range from 24--27\,°C across central and southern regions, fall below 21\,°C in the northwest and southwest, and exceed 27\,°C over the Rub al-Khali\cite{Almazroui:2012}. In contrast, the northern and interior regions experience extreme summer heat, high evaporation rates, and very low annual rainfall, severely restricting the establishment of persistent vegetation.

Atmospheric circulation further modulates regional climate variability. During the warmer months (May--October), interactions between monsoon and desert systems produce wide diurnal temperature ranges. In cooler months (November--April), subtropical jet streams and the influence of nearby water bodies moderate temperature fluctuations. Localized convergence zones, formed by the interaction of warm, moist coastal air with cooler, drier continental air, can occasionally trigger heavy rainfall and thunderstorms, particularly in mountainous and coastal transition zones\cite{UNESCWA:2023}. In summary, there are distinct climatic fingerprints for all selected land restoration candidate locations.

\begin{figure}[tbp]
\centering
\includegraphics[width=\textwidth]{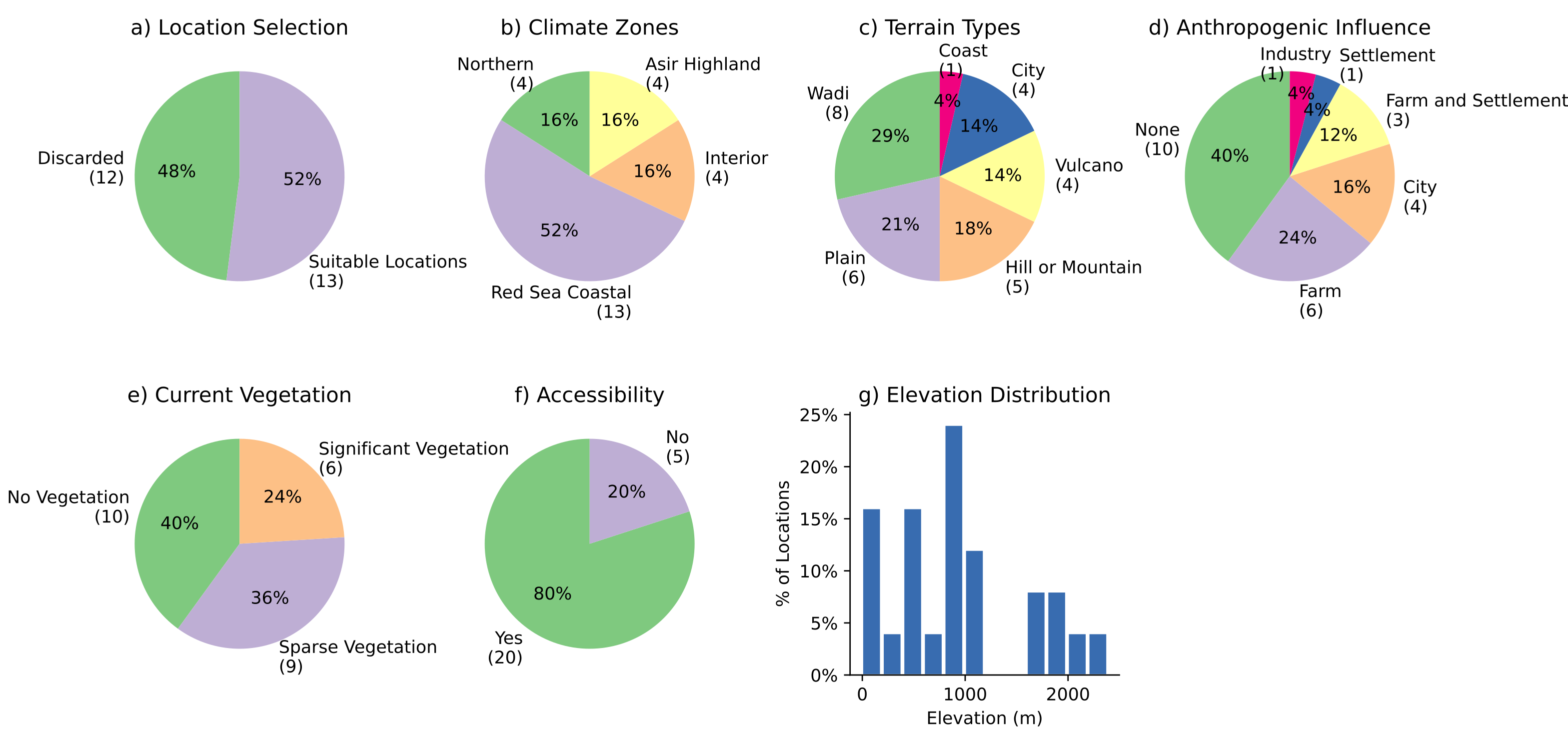}
\caption{\textbf{Analysis of the 25 selected candidate land restoration locations.} (a)~Location selection, (b)~climate zones according to Almazroui et al.\ (2015), (c)~terrain types, (d)~anthropogenic influence, (e)~current vegetation, (f)~accessibility, (g)~elevation distribution.}
\label{fig:location_stats}
\end{figure}

Most sites are located within the Red Sea coastal climate zone (Zone~2); the other sites are evenly distributed in the interior (Zone~3), the northern (Zone~1) and Asir Highland (Zone~4) zones (Figure~\ref{fig:location_stats}b). Across terrain types, one third of the sites are located within or near wadis, 21\% on plains, 18\% in hilly or mountainous terrain, the rest is located on volcanic substrates and in the middle of cities (Figure~\ref{fig:location_stats}c). Elevations range from sea level to 2,239\,m, with the majority situated below 1000\,m (Figure~\ref{fig:location_stats}g). 40\% of the predicted locations are unaffected by anthropogenic influences, and 20\% are proximal to or within settlements or cities, 24\% are close to farms indicating potential anthropogenic influence (Figure~\ref{fig:location_stats}d). Apart from the barren land, 36\% of sites exhibit sparse vegetation, 24\% show significant vegetation cover (Figure~\ref{fig:location_stats}e). Even when vegetation is present, the NDVI is below 0.1 which generally indicates barren land or rock. As discussed beforehand, the vegetation in Saudi Arabia is often concentrated in wadis surrounded by rock and barren land. This reduces the average NDVI of a 1\,km\,$\times$\,1\,km pixel while in higher resolution the NDVI within the wadi can reach 0.2 when vegetation is present, e.g., groups of perennial vegetation which can be identified using high-resolution imagery from Google Earth (Figure~\ref{fig:native_ecosystems}d and~e for a land restoration candidate location).

This second-stage screening substantially reduced the number of plausible intervention sites, demonstrating the value of climate-based pre-screening as a cost-effective front end to more detailed ecological assessment.

From an initial set of high-ranking opportunities, thirteen locations were retained as feasible restoration candidates after excluding sites constrained by urban encroachment, industrial land use or poor accessibility (Figure~\ref{fig:location_stats}d and~f). These remaining sites spanned multiple climatic zones and landforms, indicating that restoration potential is not restricted to a single ecological region but distributed across distinct dryland settings.

\subsection{Climatic Analog Ecosystems Provide Realistic Restoration Targets}

Successful restoration requires not only site selection but also realistic expectations for vegetation structure and species composition.

\begin{figure}[tbp]
\centering
\includegraphics[width=\textwidth]{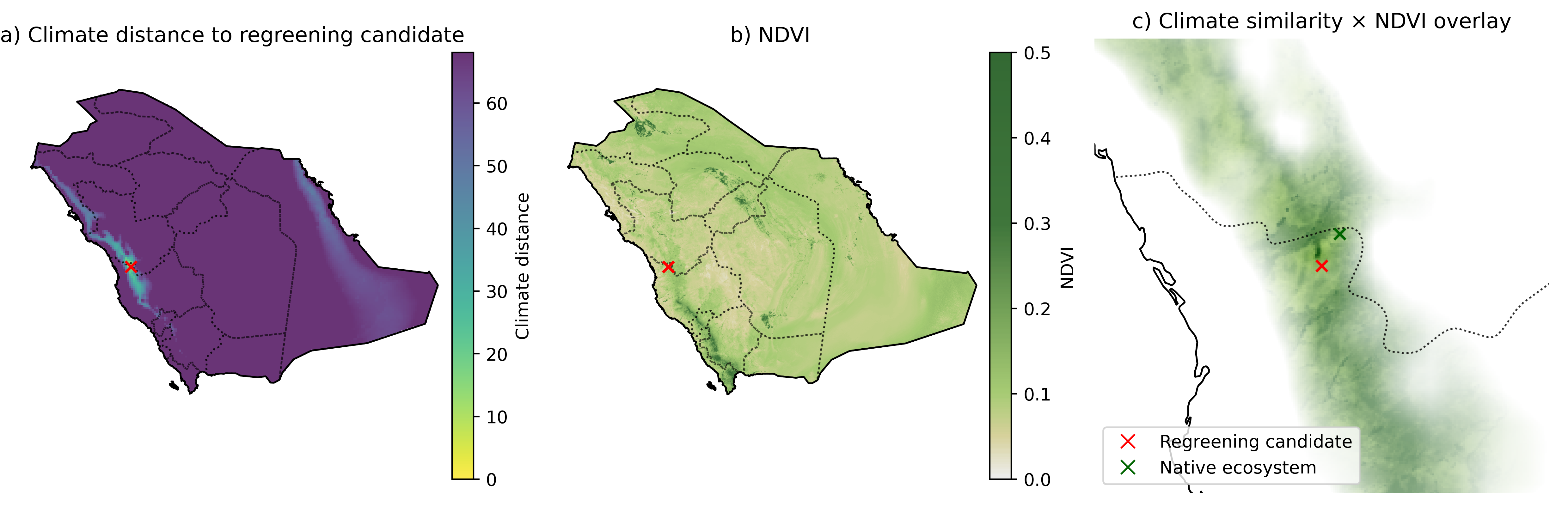}

\vspace{0.8em}

\includegraphics[width=\textwidth]{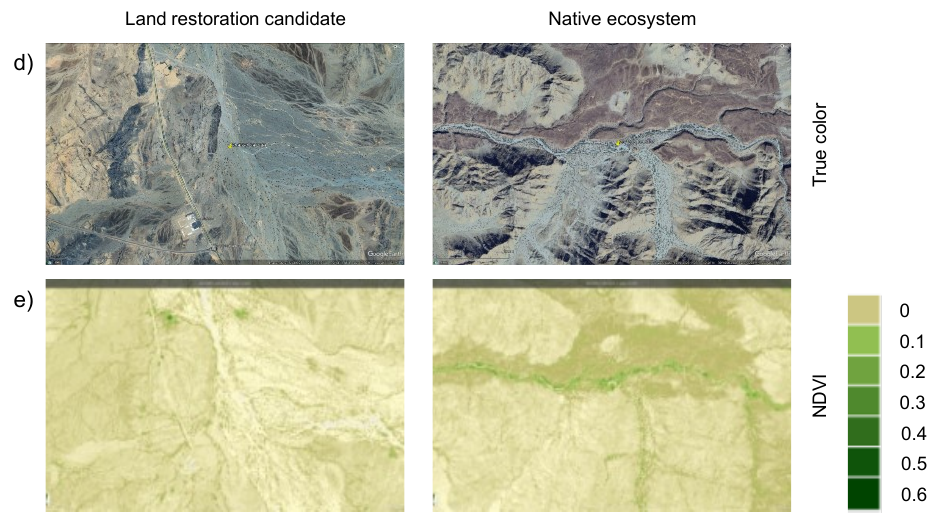}
\caption{\textbf{Native ecosystem selection process.} (a)~Climate distance map. (b)~NDVI map. (c)~Climate distance map $\times$ NDVI overlay. (d)~True color image (modified from Google Earth Airbus, 12-10-2023 for land restoration candidate location and 05-03-2023 for native ecosystem). (e)~NDVI map (contains modified Copernicus Sentinel data [04-07-2024]) for land restoration candidate location and native ecosystem. All images correspond to location~9.}
\label{fig:native_ecosystems}
\end{figure}

For each location, we use their distinct climatic fingerprint to find similar locations in a climate distance map (Figure~\ref{fig:native_ecosystems}a) exhibiting a higher NDVI value (Figure~\ref{fig:native_ecosystems}b). Then, high-resolution satellite images are used to filter out agricultural areas which have a high NDVI. Only locations with the highest summer NDVI values where perennial natural vegetation is present were chosen as a representative native ecosystem. Comparing the respective NDVI values then gives us an estimation of the land restoration potential of each individual candidate location, as the higher NDVI in native ecosystems indicates a higher vegetation density. These reference ecosystems serve as practical templates for restoration design because they represent vegetation states already sustained under similar climatic regimes.

\begin{figure}[tbp]
\centering
\includegraphics[width=\textwidth]{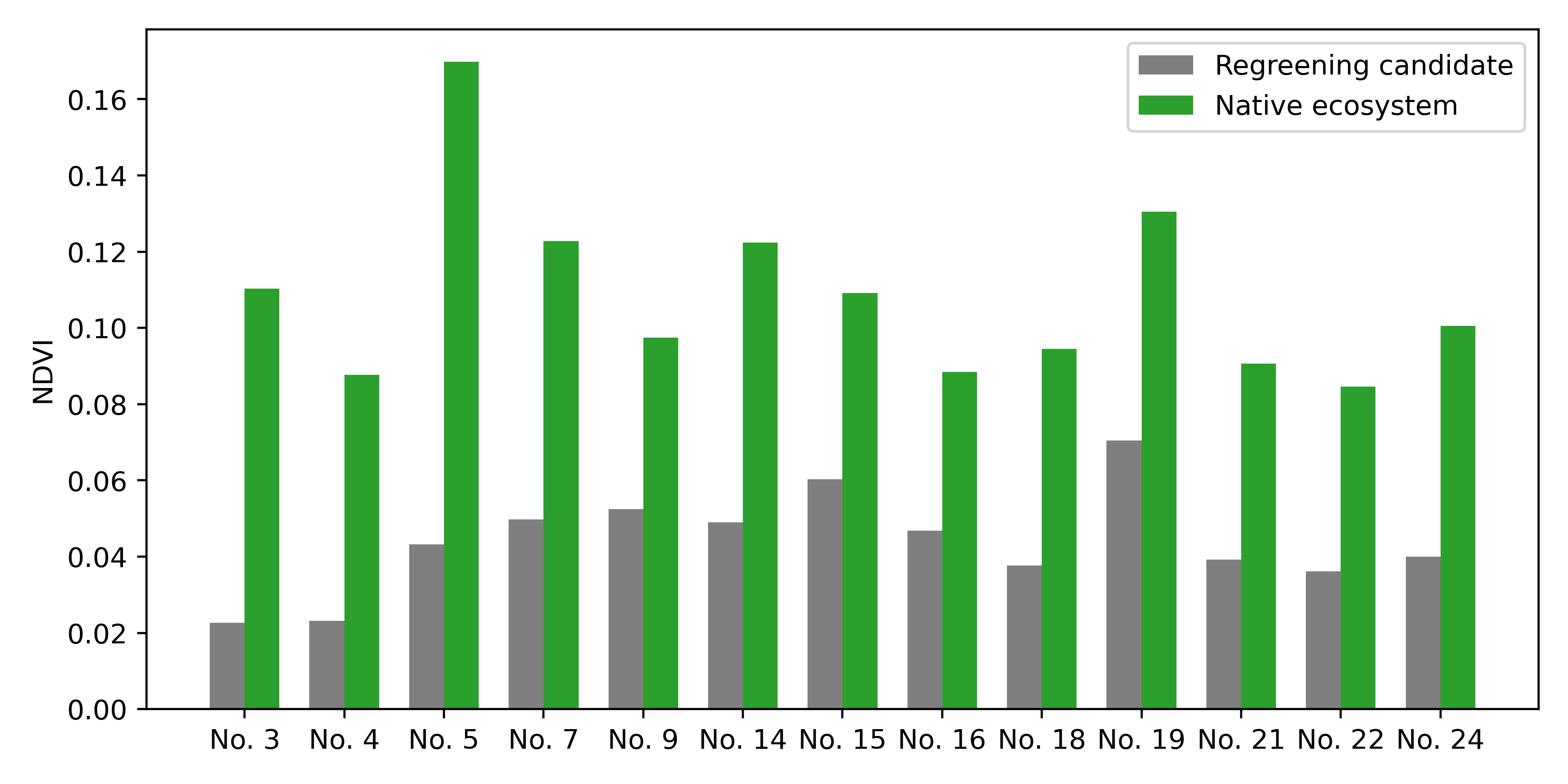}
\caption{\textbf{NDVI comparison for land restoration candidate locations and corresponding native ecosystems with climatic similarity.} Grey: Land restoration locations; green: native ecosystems. The numbers correspond to the 13 selected locations from Table~\ref{tab:candidate_locations}.}
\label{fig:ndvi_comparison}
\end{figure}

We were able to find a suitable native ecosystem as template region for each of our climatically diverse land restoration candidate locations (Figure~\ref{fig:ndvi_comparison}, Table~\ref{tab:matching_points}). Overall, the NDVI could be increased by a factor of $\sim$2.5 on average (Figure~\ref{fig:ndvi_comparison}). This suggests unrealized ecological potential under current climate conditions. The analogs can inform locally adapted species selection, target vegetation density and achievable restoration trajectories while reducing the risk of unrealistic planting goals in water-limited environments. However, further validation using on-site analysis of soil type, soil structure and water quality should be conducted to guarantee similarity to the land restoration candidate location before committing to final selection of both land restoration location and corresponding native ecosystem.

\section{Discussion}

Regreening projects in arid regions often rely on sustained human intervention (particularly irrigation) to ensure the survival of newly planted vegetation. Because this irrigation water is typically drawn from aquifers that are replenished by precipitation far slower than the demands of agriculture and population growth, such projects can place additional strain on already limited groundwater resources\cite{Fu:2025}. Alternatives, such as recycled sewage water or temporary rainwater reservoirs, are equally limited, making irrigation unsustainable long term. Additionally, the newly planted vegetation will adapt to higher soil moisture, which will make the plants more vulnerable to drought stress once irrigation is eventually stopped. An ill-considered trade-off between land restoration efforts and resource availability will inevitably exacerbate desertification. Therefore, finding locations that offer a supportive environment is crucial before starting any regreening programs. In this study, we aim to develop a framework to prioritize restoration opportunities for field validation.

In previous work, similar problems have been addressed. Roebroek et al.\ estimate a worldwide potential for tree growth under natural conditions\cite{Roebroek:2025}. Piri Sahragard et al.\ evaluate the suitability of habitats for almond growth in southern Iran\cite{PiriSahragard:2021}. Dakhil et al.\ focus on the importance of clay content in soil for the prediction of juniper growth in the Mediterranean region\cite{Dakhil:2022}. MacKenzie et al.\ extrapolate future climate conditions and their impact on tree growth in British Columbia\cite{MacKenzie:2021}. Other regions, such as the Scottish Highlands and southeast Asia, have been addressed in similar research as well\cite{Goodall:2026,Tiansawat:2022}.

A common, potential limitation of most of the related works is that only a relatively small number of variables are included in the analysis, for example 7 climate variables (of which 4 are different statistical measures of temperature) and 6 soil variables\cite{Dakhil:2022} or 19 climate variables, which are all statistical measures either derived from temperature or precipitation\cite{Tiansawat:2022}.

Here, we attempt to maximize classification accuracy by including all available climate data, resulting in 23 individual variables, densely sampled on a 3-hour basis over 5 years. This means that any statistical measure (such as daily or monthly averages or minimal and maximal variables) of each of these variables is also available to the machine learning models. The input compression methods we use (selection of dominant frequencies via Fourier transformation and latent vector compression through autoencoder) are designed to leave these largely intact. Avoiding a preselection of intuitively promising variables avoids human selection bias.

Various types of machine learning models for vegetation growth analysis have been proposed in the literature, including support vector machines\cite{Goodall:2026}, generalized linear models\cite{PiriSahragard:2021,Dakhil:2022}, or multilayer perceptrons\cite{Goodall:2026}, with maximum entropy\cite{PiriSahragard:2021,Goodall:2026,Tiansawat:2022} and random forest\cite{Fu:2025,Roebroek:2025,Dakhil:2022,Goodall:2026} being the most popular. Given our large number of input variables ($\sim$380,000 per location before compression), we opted for the BLUP model due to its proven robustness in similar problems in genetics, and for a neural network due to their performance in detecting complex patterns\cite{Habier:2007,Dsouza:2020}.

Even though the two evaluated machine learning models are structurally different, both identify the same most suitable locations. This is a strong indicator that the results are consistent and the chosen ML-based approach is suitable for the given task. The evaluation of the newly introduced CSS values is hindered by the lack of ground-truth values. However, the plausibility and usefulness of our predictions are shown through two important analyses:

Even though we use binary input labels, the predictions are more continuous after reclassifications. Evaluation revealed that the newly assigned values are consistent with the extent of present vegetation, indicating that more suitable locations generally also yield higher NDVI values. This suggests that locations with a high CSS prediction, but a low NDVI have a good potential for land restoration.

When applying our models across Saudi Arabia, the resulting patterns are meaningful. Larger areas with high vegetation cover, such as the Jazan region or the Asir mountains, are identified as high-potential areas, while drier parts, such as the Empty Quarter in the southeast, are consistently predicted to have low values. The predictions are therefore consistent with expectations\cite{Ghazal:2015,Valjarevic:2023,AlmalkiK:2022}. On a finer scale, we find that the agricultural regions, which exhibit very dense coverage of human-supported vegetation, are still classified as low-potential regions because they rely heavily on additional irrigation. On the other hand, a larger region in the Hail province, known from the literature to be degraded\cite{AlRowaily:2018}, is predicted to support higher vegetation cover than is currently present (Figure~\ref{fig:opportunity_map}c).

With these results, the CSS can be used for the climatic preselection of suitable land restoration locations. However, the identified 25 candidate locations needed to be analyzed further to address the other limiting criteria that are not covered by our training data.

One of the most important factors is soil quality. For example, many of the selected locations fall into volcanic areas, which typically have soil composed of coarse basalt, which does not support the extensive root systems of plants\cite{MunozSabater:2021,Burbano:2022}. However, at the edge of the volcanic area, eroded basalt mixes with sandy soil, increasing support for plant growth.

Anthropogenic pressures, particularly urban expansion or industrial areas, represent a competing factor to regreening efforts. Rapid urbanization in the coastal Red Sea region and the Asir Highlands has already been linked to reductions in natural vegetation cover\cite{Moatamed:2021}. These pressures must be carefully considered when planning restoration or conservation initiatives to ensure long-term ecological sustainability. The locations from our final selection are accessible by road but do not compete with settlement expansions.

Ecosystem degradation is measured in different levels, from degraded native ecosystem to highly modified degraded ecosystem\cite{LiuJ:2024}. Depending on the level of degradation, stepwise restorative measures must be taken to achieve a fully restored native ecosystem\cite{LiuJ:2024}. With increasing degradation, the restoration efforts become increasingly difficult and costly\cite{LiuJ:2024}. In many of our predicted locations, vegetation was still present. The existing vegetation indicates a level of degradation that can be restored in the future by expanding the current ecosystem.

To define realistic restoration targets using adapted perennial vegetation, native ecosystems can function as templates for plant selection as well as achievable vegetation density. We identified climatically analogous reference ecosystems with higher NDVI that represent denser native vegetation. Next to climate factors, other factors also influence vegetation density, e.g. soil type, soil structure and water quality. An on-site analysis of native ecosystems would have to be conducted to ensure physical and chemical similarity to the selected land restoration candidate location.

Emulating vegetation cover from intact native ecosystems in similar regions will avoid overplanting, which ensures that limited resources are not overstretched.

Limitations of our work include the relatively low resolution of the ERA5 dataset (9\,$\times$\,9\,km) and the global NDVI dataset (1\,$\times$\,1\,km) used. For example, the densest vegetation in Saudi Arabia is commonly found in relatively narrow wadis, while the surrounding area is often barren. This can impact the prediction validity and necessitates future refinement of the exact land restoration locations, as our suggested locations are only precise up to the resolution of the climate dataset.

The limited sample size of the training dataset increases overfitting and limits the training accuracy. Additionally, the ERA5 dataset is known to have limited accuracy in Saudi Arabia due to the lack of a dense network of weather stations\cite{LiuR:2024}. These uncertainties introduce noise to the training data, further limiting the model fitting accuracy. On the other hand, the input labels lack expressiveness, since the CSS value does not directly correspond to any physical measurement, which necessitated binary input labels. Validity despite these shortcomings is mostly demonstrated by downstream analysis, showing the overall meaningfulness of the classification results. Finally, while the framework is transferable to other regions, new training data for the models should be collected because microclimates vary across countries.

The crucial factor for successful regreening projects is the correct selection of suitable locations and plant species. As country-wide detailed analysis is prohibitively costly, an efficient preselection step is required. By adjusting suitability expectations, both broad regions and specific candidate locations can be identified.

Apart from Saudi Arabia and its \emph{Saudi Green Initiative}, our framework can be applied to similar dryland regions, including the Arabian Peninsula, North Africa, Central Asia, Australia and parts of the southwestern United States. Its relevance is particularly pronounced in resource-constrained regions---such as many countries in Africa---where land degradation is severe and restoration efforts depend heavily on external funding, making efficient and evidence-based prioritization essential. Our framework helps define realistic restoration targets and expectations for policy makers, scientists and stakeholders, reducing the risk of failure and improving long-term outcomes.

\section{Methods}

\subsection{Study Area}

The spatial extent is from 12° to 34° in latitude and 34° to 56° in longitude, covering a quadratic area around Saudi Arabia (Figure~\ref{fig:sample_positions}).

\subsection{Data Sources}

We rely on the publicly available ERA5-Land dataset as the primary source of climatic information\cite{MunozSabater:2021}. The ERA5-Land dataset, developed by the European Centre for Medium-Range Weather Forecasts (ECMWF) under the framework of the Copernicus Climate Change Service Programme, provides high-quality reanalysis data for various climate variables. Of all available variables, we used the 23 most relevant ones, listed in Table~\ref{tab:era5_variables}. We excluded parameters that play only a very minor role in plant growth or in our region of interest, such as lake temperature or snow coverage. The complete list and documentation of all available parameters in the ERA5 dataset are found at \url{https://confluence.ecmwf.int/display/CKB/ERA5-Land}. The ERA5-Land is a reanalysis database with a consistent view of the evolution of land variables for several decades at an enhanced spatial resolution of 9\,km (0.08°). For a detailed description of the ERA5-Land dataset, the reader is referred to Mu\~noz-Sabater et al.\cite{MunozSabater:2021}, and the online documentation: DOI: \href{https://doi.org/10.24381/cds.e2161bac}{10.24381/cds.e2161bac} (accessed 21-03-2025). We analyzed the 5-year period from January 2020 to December 2024, using data points every 3 hours (0:00, 3:00, 6:00, 9:00, 12:00, 15:00, 18:00, 21:00) each day. This corresponds to 14,616 data points per variable. The spatial extent reaches from 12° to 34° in latitude and 34° to 56° in longitude, covering a quadratic area around Saudi Arabia. Overall, our dataset therefore contains 23 variables with 221\,$\times$\,221 spatial samples and 14,616 temporal samples, i.e., around 138\,GB of uncompressed data.

For the \emph{Normalized Difference Vegetation Index} (NDVI) we used two datasets: For Saudi-wide analysis we used the \emph{eVIIRS Global NDVI}\cite{Vermote:2014}, which provides a spatial resolution of 1000\,m and a temporal resolution of 10 days. For our analysis we used the datasets from January 2020 to December 2024 (DOI: \href{https://doi.org/10.5066/P9QOEFNP}{10.5066/P9QOEFNP}, accessed 14-04-2025).

For detailed analysis of individual locations, we use the NDVI as processed in \emph{Copernicus Browser} from Sentinel-2 satellite data. It provides a 10\,m spatial and 5-day temporal resolution (\url{https://browser.dataspace.copernicus.eu/}). For both of these datasets we analyzed the years 2020--2024.

Additionally, we analyzed individual locations using the high-resolution true color satellite images and elevation information provided by \emph{Google Earth Airbus}.

\subsection{Training Data}

We selected a total of 230 sample locations for our training. The most important criterion for labelling was the five-year Sentinel-2 NDVI time series, in which persistent vegetation was used as an indicator of high suitability, and persistent absence of vegetation as low suitability. These two categories are, however, insufficient, as they correspond to actual plant growth rather than potential growth (which often, though not necessarily, aligns). We therefore also added two more categories with locations with more vegetation than naturally supported (e.g., farming regions with irrigation) and areas that should be suitable for growth but lack vegetation (e.g., degraded locations, where natural vegetation was removed through anthropogenic factors such as overgrazing or logging).

Generally, we selected sample locations based on their summer NDVI value in 2024 and then confirmed presence of vegetation using the Google Earth high-resolution true color images. To verify long-term vegetation presence, we looked at a 5-year NDVI time series from 2020 to 2024. As discussed in the introduction, the generally applied NDVI thresholds are a poor fit to arid regions. Analyzing the high-resolution images of our sample locations, we find that 0.15 is a more sensible threshold for the presence of vegetation in our target region during the whole time period.

Since we classified locations based on their climate vectors, each sample location corresponds to a single pixel in the ERA5-Land dataset, i.e., an area of 9\,km\,$\times$\,9\,km. A minimum spacing of 9\,km was enforced between locations to ensure each location corresponds to a unique pixel.

In detail, the selection criteria were:

\begin{itemize}[itemsep=2pt]
  \item \textbf{High climate suitability, high vegetation prevalence} (101 samples): Perennial natural vegetation in a significant sub-area (minimum radius of 500\,m or 2\,km length of wadi) with an average NDVI above 0.2 and no long-time drops below 0.15 over 5 years. Agricultural areas and settlements are excluded.
  \item \textbf{Low climate suitability, low vegetation prevalence} (101 samples): 5-year absence of vegetation in the whole area, where the NDVI is consistently below 0.15.
  \item \textbf{Low climate suitability, high vegetation prevalence} (14 samples): An area containing agricultural land usage but no natural perennial vegetation.
  \item \textbf{High climate suitability, low vegetation prevalence} (14 samples): Locations described in the literature as degraded through anthropogenic influences\cite{AlRowaily:2018,Salih:2021}. An area with no significant perennial vegetation, but potential seasonal coverage (e.g., grass) for fewer than 4 months.
\end{itemize}

\subsection{BLUP Analysis}

We performed an automated, frequency analysis-based reduction of the input data to improve stability and training performance. Almost all climate variables are subject to periodic changes on different time scales (hourly, daily, monthly). This naturally suggests a data analysis in the frequency domain, obtained via the Fourier transformation. Sorting the frequency phasors by amplitude is a natural way to compress data, since frequencies with a very small contribution to the overall signal can be ignored with little loss of information.

We computed a number of reduced datasets ranging from 2 to 1024 frequencies, in increments of 2$\times$. The frequencies were selected by individual contribution to each variable, i.e., the same number of frequencies is used for each variable, but each variable uses different frequencies. The frequency selection and normalization value computation were performed on data from the sample locations. This makes the computation of normalized frequency vectors only depend on the classifier's training data, not on the region where it is later evaluated.

\subsection{Neural Network Classifier}

Following common neural network best practices, we trained an autoencoder to learn an optimal feature representation for different vector lengths. These latent vectors were then used by a second network to perform the classification.

The autoencoder architecture follows the standard hourglass approach, as shown in full in Figure~\ref{fig:autoencoder}. To mitigate overfitting, we applied Gaussian noise, per-variable dropout layers, and batch normalization. We used a brute-force search to determine suitable hyperparameters for training (such as batch size and learning rate). These are shown in Table~\ref{tab:network_params}. For the autoencoder loss, we compared values in normalized frequency space rather than in the temporal domain.

The network architecture of the neural classifier is shown in Figure~\ref{fig:classifier}. Again, we make use of Gaussian noise and dropout layers to reduce overfitting. As input, we used latent vectors of sizes between 4 and 1024, in increments of 2$\times$. Rather than a sigmoid activation function that would be common for binary classification, we used a linear activation layer for a RMSE loss function. This results in a smoother distribution of the output values, which allows for a better ranking among predictions.

Note that the input sizes for BLUP and the neural network are not directly comparable. While the Fourier coefficients are counted per variable (23 in total), the latent vector dimensionality is the overall size. However, each latent vector dimension carries much more information due to the autoencoder's efficient compression.

Full specification of the training network setup and training process is documented via the accompanying source code.

\subsection{Prediction Analysis}

The prediction values of both the BLUP and neural network methods are mostly contained in the interval $[0, 1]$, where occasionally values less than 0 or greater than 1 can be produced. NDVI values are always in the interval $[0, 1]$. Although their ranges are very similar, the prediction value and NDVI cannot be compared directly because their units differ.

There is therefore no a-priori relationship between prediction and NDVI values. To estimate the most meaningful mapping, we reclassify the input locations. This gives us new, continuous labels in $[0, 1]$ and corresponding NDVI values. We then fit a linear function to obtain the most meaningful mapping between them. Note that only the ``high suitability, high prevalence'' and the ``low suitability, low prevalence'' classes are used for this fit, as their CSS values are expected to correspond closely to the NDVI values. The mappings are shown in Figure~\ref{fig:reclassification}. In Figure~\ref{fig:opportunity_map}, the adjusted predictions are then compared against the NDVI values.

We only use NDVI data for the summer months to avoid capturing seasonal vegetation. Specifically, we select the NDVI measurements captured between day 80 and day 256 (corresponding to dates between mid-March and mid-September) for each of the 5 years and average them.

\subsection{Selection of Native Ecosystems Corresponding to Land Restoration Candidate Locations}

In order to find a suitable selection of native ecosystems as templates for specific regreening locations, we search for nearby locations with a similar climate and a denser vegetation cover.

We define a similarity metric between climate vectors through the L2 norm of the truncated Fourier coefficients, using the first 32 channels. Then, taking a potential regreening candidate location as the reference location, we compute climate distances for the whole target region (Figure~\ref{fig:native_ecosystems}a). We find a strong correspondence between spatial distance and climate similarity; however, locations further away can also have reasonably similar climatic conditions, as seen in the eastern region in Figure~\ref{fig:native_ecosystems}a.

Next, we overlay the NDVI (Figure~\ref{fig:native_ecosystems}b) with this climate difference map and use this as a preselection to find locations that have already established plant growth and are climatically similar to our selected location.

We use high-resolution true color satellite images to verify candidates for native ecosystems, e.g., to exclude agricultural usage, locations in settlements, or locations with obviously different landforms (sandy plains vs.\ rocky hills). With this, we end up with a nearby reference location that should be further studied on-site for their vegetation composition in order to choose adapted plant species for regreening the candidate location.

\subsection{Data and Code Availability}

Evaluation data and analysis code is published in the supplementary material.

\section*{Acknowledgments and Funding}

We thank Aleksandar Cvejic for helpful discussions. This work was funded by the KAUST baseline funding.

\clearpage

\setcounter{figure}{0}
\renewcommand{\thefigure}{S\arabic{figure}}
\setcounter{table}{0}
\renewcommand{\thetable}{S\arabic{table}}

\section*{Supplementary Material}

Additional data is provided in the accompanying ZIP-archive:
\begin{itemize}
  \item Sample Locations: The 230 sample locations
  \item Prediction maps: Individual maps for BLUP and neural network
  \item Source code, including the exact definitions of the network architecture
\end{itemize}

\begin{figure}[tbp]
\centering
\includegraphics[width=0.88\textwidth]{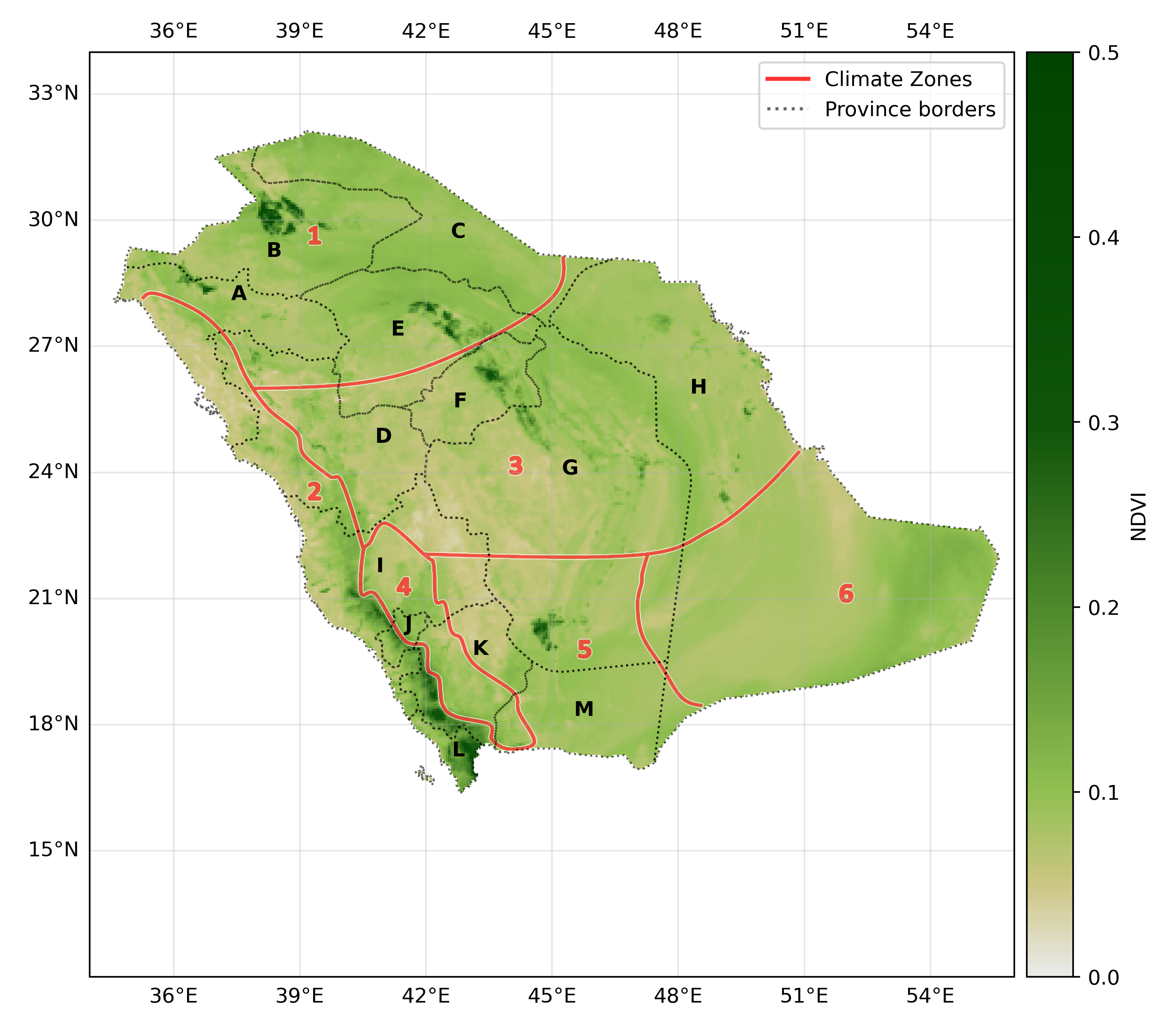}
\caption{\textbf{Map of Saudi Arabia.} NDVI, provincial borders and climate zones are shown. Provinces: A~Tabuk, B~Al Jawf, C~Northern Border, D~Madinah, E~Hail, F~Qassim, G~Riyadh, H~Eastern Province, I~Makkah, J~Baha, K~Asir, L~Jazan, M~Najran. Climate regions: 1~Northern, 2~Coastal Red Sea, 3~Interior, 4~Asir Highland, 5~South-Western, and 6~South Eastern Region, adapted from Almazroui et al.\ (2015).}
\label{fig:saudi_ndvi}
\end{figure}

\begin{table}[tbp]
\centering
\small
\caption{\textbf{Selected climate variables from the ERA5 dataset.}}
\label{tab:era5_variables}
\begin{tabular}{rlrl}
\toprule
Index & Name & Unit & Description \\
\midrule
1  & \texttt{d2m}    & K                       & 2m dewpoint temperature \\
2  & \texttt{evabs}  & m of water equivalent   & Evaporation from bare soil \\
3  & \texttt{evaow}  & m of water equivalent   & Evaporation from open water surfaces excl.\ oceans \\
4  & \texttt{evatc}  & m of water equivalent   & Evaporation from the top of canopy \\
5  & \texttt{evavt}  & m of water equivalent   & Evaporation from vegetation transpiration \\
6  & \texttt{sp}     & Pa                      & Surface pressure \\
7  & \texttt{src}    & m of water equivalent   & Skin reservoir content \\
8  & \texttt{sro}    & kg/m\textsuperscript{2} & Surface runoff \\
9  & \texttt{ssrd}   & J/m\textsuperscript{2}  & Surface solar radiation downwards \\
10 & \texttt{ssro}   & m                       & Sub-surface runoff \\
11 & \texttt{stl1}   & K                       & Soil temperature level 1 (0--7\,cm) \\
12 & \texttt{stl2}   & K                       & Soil temperature level 2 (7--28\,cm) \\
13 & \texttt{stl3}   & K                       & Soil temperature level 3 (28--100\,cm) \\
14 & \texttt{stl4}   & K                       & Soil temperature level 4 (100--289\,cm) \\
15 & \texttt{strd}   & J/m\textsuperscript{2}  & Surface thermal radiation downwards \\
16 & \texttt{swvl1}  & m\textsuperscript{3}/m\textsuperscript{3} & Volumetric soil water layer 1 (0--7\,cm) \\
17 & \texttt{swvl2}  & m\textsuperscript{3}/m\textsuperscript{3} & Volumetric soil water layer 2 (7--28\,cm) \\
18 & \texttt{swvl3}  & m\textsuperscript{3}/m\textsuperscript{3} & Volumetric soil water layer 3 (28--100\,cm) \\
19 & \texttt{swvl4}  & m\textsuperscript{3}/m\textsuperscript{3} & Volumetric soil water layer 4 (100--289\,cm) \\
20 & \texttt{t2m}    & K                       & 2m temperature \\
21 & \texttt{tp}     & m                       & Total precipitation \\
22 & \texttt{u10}    & m/s                     & 10m u component of wind \\
23 & \texttt{v10}    & m/s                     & 10m v component of wind \\
\bottomrule
\end{tabular}
\end{table}

\begin{figure}[tbp]
\centering
\begin{minipage}{0.48\textwidth}
  \centering
  \includegraphics[width=\textwidth]{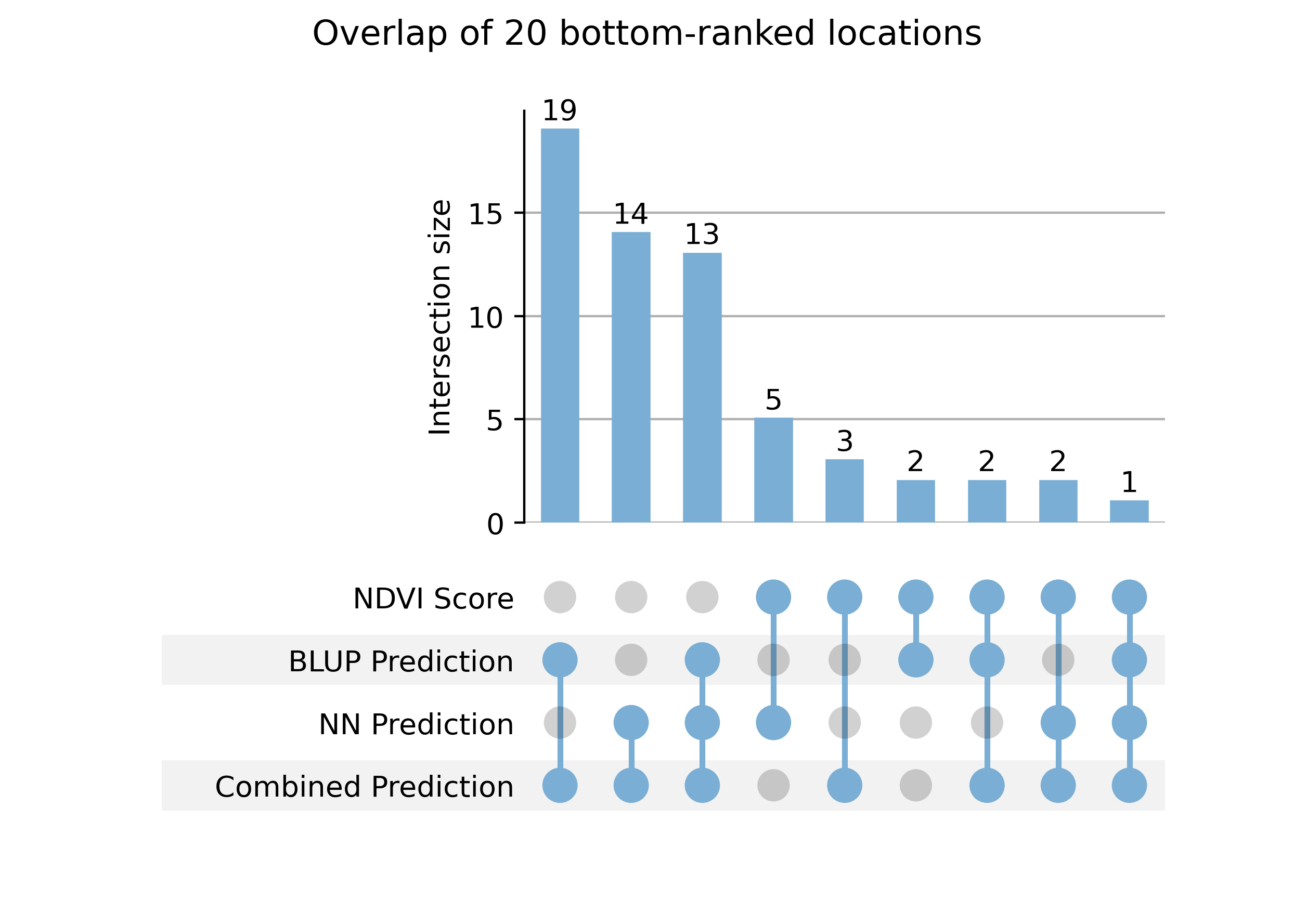}
\end{minipage}
\hfill
\begin{minipage}{0.48\textwidth}
  \centering
  \includegraphics[width=\textwidth]{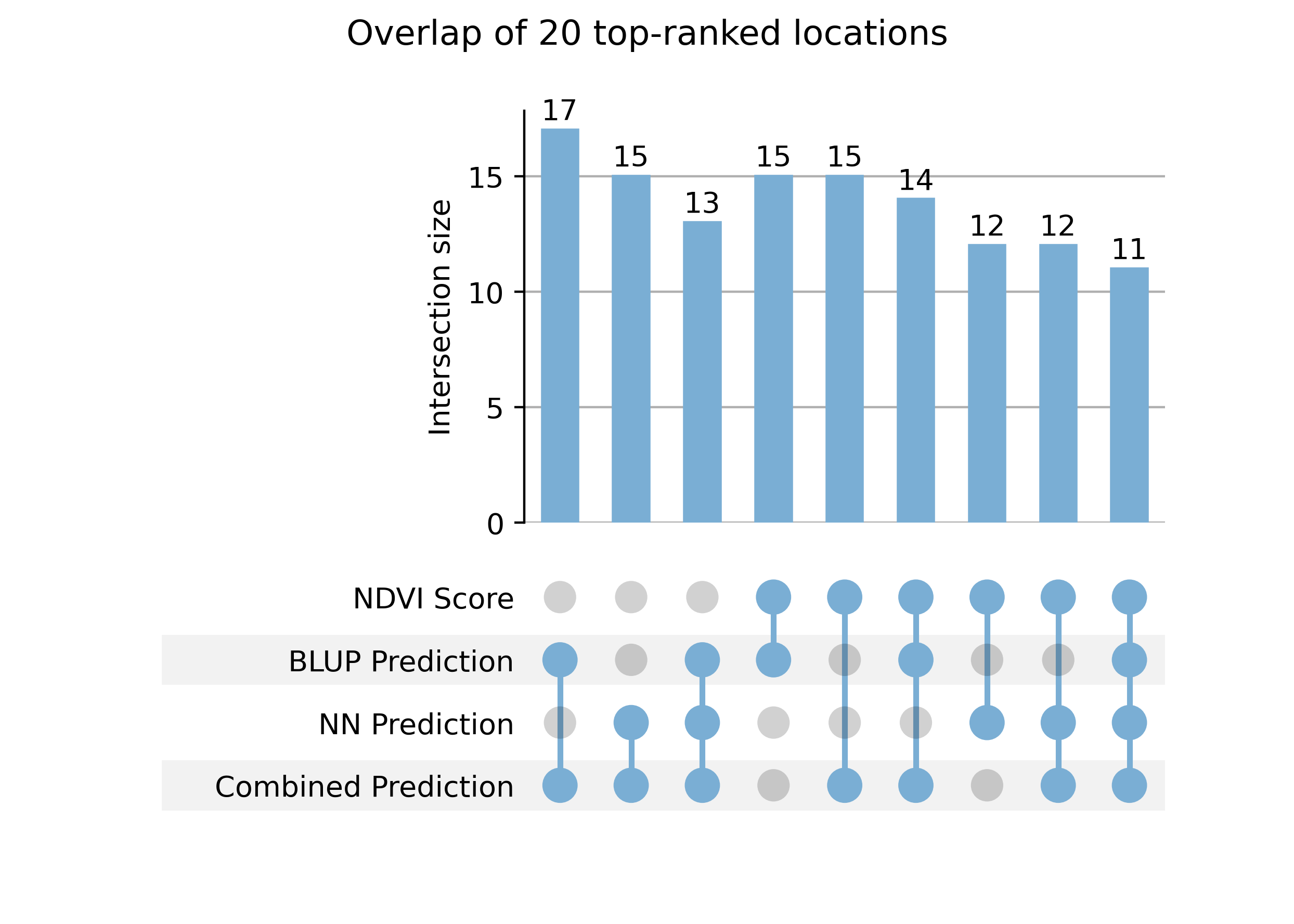}
\end{minipage}
\caption{\textbf{UpSet plot of the 20 highest and lowest ranked sample locations after reclassification.} The analysis was performed according to the description in Section~2.1. All 101 vegetated sample locations were ranked according to the 4 categories (NDVI value, reclassification value after training the BLUP, neural network (NN), and combined prediction) individually, and the 4 sets of 20 top- and bottom-ranked locations were extracted. For each combination of these 4 sets (bottom), the number of elements of their intersection is computed and shown in the bar plot (top).}
\label{fig:overlap}
\end{figure}

\begin{figure}[tbp]
\centering
\includegraphics[width=0.8\textwidth]{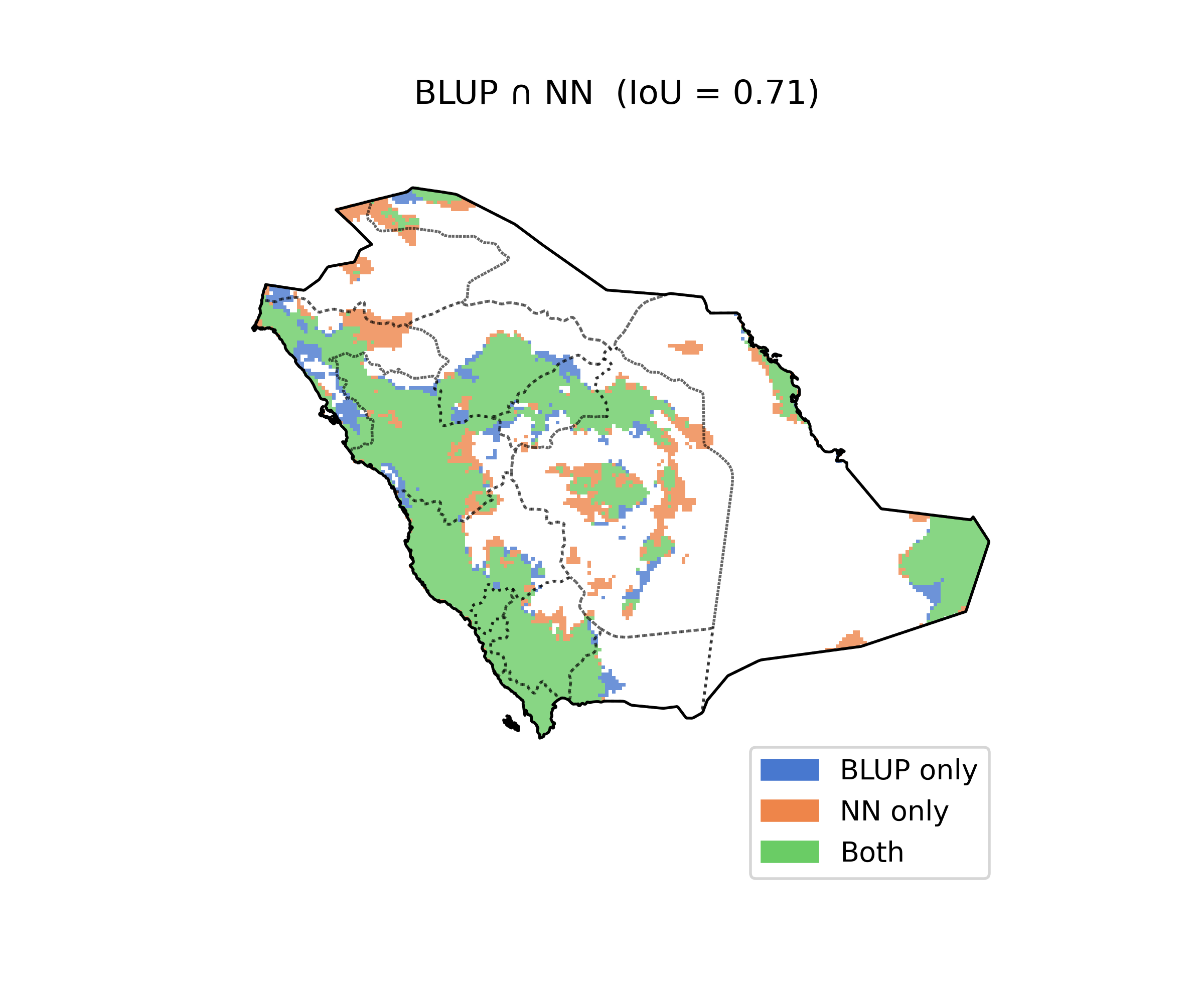}
\caption{\textbf{CSS overlap between BLUP and neural network predictions.} Agreement between the BLUP and neural network CSS after thresholding the CSS value at 0.5, measured by the Intersection over Union (IoU) score of positive locations.}
\label{fig:iou}
\end{figure}

\begin{landscape}
\begin{table}[p]
\centering
\small
\caption{\textbf{Best locations from comparison between CSS and NDVI.} The 13 selected locations are highlighted in green. They are shown in Figure~\ref{fig:opportunity_map}c.}
\label{tab:candidate_locations}
\begin{tabular}{rllllrrll}
\toprule
No. & CSS & Province & Climate Zone & Terrain & Elevation (m) & Vegetation & Anthropogenic Influence & Accessibility \\
\midrule
1 & 0.912 & Tabuk & Red Sea Coastal & Volcano & 1149 & No Vegetation & None & No \\
2 & 0.806 & Madinah & Red Sea Coastal & Hill or Mountain, Wadi & 837 & Sparse Vegetation & None & No \\
\rowcolor{rowColorGood}3 & 0.825 & Madinah & Red Sea Coastal & Wadi & 526 & Sparse Vegetation & None & Yes \\
\rowcolor{rowColorGood}4 & 0.846 & Madinah & Red Sea Coastal & Wadi & 646 & Significant Vegetation & Farm & Yes \\
\rowcolor{rowColorGood}5 & 0.942 & Madinah & Red Sea Coastal & Wadi & 386 & Significant Vegetation & Farm & Yes \\
6 & 0.835 & Madinah & Interior & Volcano & 833 & No Vegetation & Farm and Settlement & No \\
\rowcolor{rowColorGood}7 & 0.970 & Makkah & Red Sea Coastal & Wadi & 421 & Significant Vegetation & None & Yes \\
8 & 0.898 & Makkah & Red Sea Coastal & Plain & 169 & No Vegetation & Farm and Settlement & Yes \\
\rowcolor{rowColorGood}9 & 1.025 & Makkah & Red Sea Coastal & Wadi & 589 & Significant Vegetation & None & Yes \\
10 & 0.838 & Madinah & Red Sea Coastal & Volcano & 850 & No Vegetation & Farm and Settlement & Yes \\
11 & 0.916 & Makkah & Red Sea Coastal & City & 402 & No Vegetation & City & Yes \\
12 & 1.033 & Hail & Interior & Volcano & 1732 & No Vegetation & None & No \\
13 & 1.002 & Makkah & Red Sea Coastal & City & 1683 & No Vegetation & City & Yes \\
\rowcolor{rowColorGood}14 & 0.955 & Makkah & Red Sea Coastal & Plain & 80 & Significant Vegetation & None & Yes \\
\rowcolor{rowColorGood}15 & 1.018 & Makkah & Red Sea Coastal & Plain & 33 & Sparse Vegetation & Farm & Yes \\
\rowcolor{rowColorGood}16 & 0.919 & Hail & Northern & Hill or Mountain, Wadi & 1141 & Sparse Vegetation & None & Yes \\
17 & 0.870 & Hail & Northern & City & 987 & No Vegetation & City & Yes \\
\rowcolor{rowColorGood}18 & 0.931 & Hail & Northern & Plain & 1012 & No Vegetation & Farm & Yes \\
\rowcolor{rowColorGood}19 & 1.064 & Aseer & Asir Highland & Hill or Mountain & 2046 & Significant Vegetation & Farm & Yes \\
20 & 1.032 & Aseer & Asir Highland & Hill or Mountain & 2239 & Sparse Vegetation & None & No \\
\rowcolor{rowColorGood}21 & 0.869 & Qassim & Interior & Plain & 973 & Sparse Vegetation & Farm & Yes \\
\rowcolor{rowColorGood}22 & 0.917 & Hail & Northern & Plain & 893 & Sparse Vegetation & None & Yes \\
23 & 0.984 & Aseer & Asir Highland & City & 1970 & No Vegetation & City & Yes \\
\rowcolor{rowColorGood}24 & 0.958 & Najran & Asir Highland & Hill or Mountain, Wadi & 1841 & Sparse Vegetation & Settlement & Yes \\
25 & 0.869 & Eastern Province & Interior & Coast & 4 & Sparse Vegetation & Industry & Yes \\
\bottomrule
\end{tabular}
\end{table}
\end{landscape}

\begin{landscape}
\begin{table}[p]
\centering
\small
\caption{\textbf{Predicted land restoration locations vs.\ native ecosystem correspondences.}}
\label{tab:matching_points}
\begin{tabular}{rllrrrr}
\toprule
No. & Selected Location & Intact Ecosystem Location & Pred.\ NDVI & Int.\ NDVI & Climate Dist. & Spatial Dist.\ (km) \\
\midrule
3  & 25°59'35.9"N 38°00'42.5"E & 25°53'55.7"N 38°02'13.6"E & 0.0226 & 0.1103 & 13.0867 & 10.8 \\
4  & 25°17'17.9"N 38°09'56.5"E & 25°15'08.6"N 38°13'49.8"E & 0.0231 & 0.0877 & 0.0000 & 7.6 \\
5  & 24°39'59.8"N 38°19'49.8"E & 24°35'39.5"N 38°16'05.9"E & 0.0432 & 0.1698 & 30.6510 & 10.2 \\
7  & 22°44'28.0"N 39°36'13.7"E & 22°40'21.4"N 39°36'07.2"E & 0.0497 & 0.1228 & 0.0000 & 7.6 \\
9  & 22°57'07.9"N 39°44'20.0"E & 23°07'15.2"N 39°49'59.9"E & 0.0525 & 0.0974 & 17.4009 & 21.1 \\
14 & 19°23'33.0"N 41°12'59.4"E & 19°24'29.2"N 41°11'06.7"E & 0.0490 & 0.1224 & 0.0000 & 3.7 \\
15 & 19°08'22.6"N 41°13'30.0"E & 18°36'23.0"N 41°31'59.5"E & 0.0602 & 0.1091 & 32.0678 & 67.6 \\
16 & 27°09'43.2"N 41°37'50.5"E & 26°57'25.7"N 41°48'44.3"E & 0.0468 & 0.0885 & 10.4240 & 29.0 \\
18 & 26°51'41.8"N 42°15'30.2"E & 27°01'46.9"N 42°08'55.7"E & 0.0377 & 0.0945 & 13.6228 & 21.6 \\
19 & 18°43'32.9"N 42°24'12.6"E & 18°46'17.4"N 42°23'29.8"E & 0.0705 & 0.1304 & 18.6523 & 5.2 \\
21 & 26°18'06.1"N 42°30'36.4"E & 26°24'37.4"N 42°39'12.2"E & 0.0392 & 0.0907 & 10.4677 & 18.7 \\
22 & 27°02'40.9"N 42°31'08.0"E & 27°05'44.5"N 42°25'52.7"E & 0.0361 & 0.0846 & 6.6562 & 10.4 \\
24 & 18°16'16.0"N 43°52'14.2"E & 18°12'00.4"N 43°56'12.1"E & 0.0400 & 0.1005 & 15.1577 & 10.5 \\
\bottomrule
\end{tabular}
\end{table}
\end{landscape}

\begin{figure}[tbp]
\centering
\includegraphics[height=0.7\textheight]{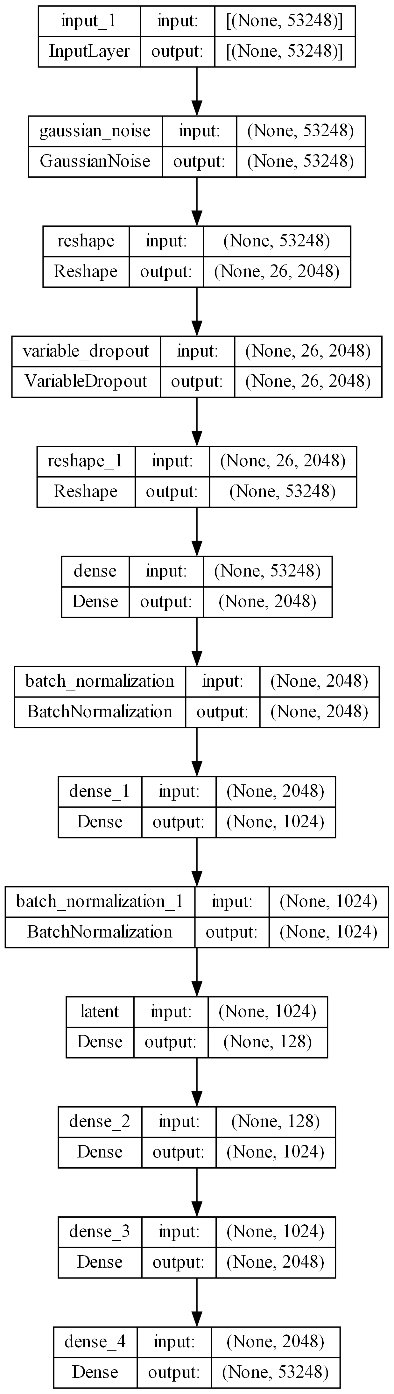}
\caption{\textbf{Autoencoder architecture.}}
\label{fig:autoencoder}
\end{figure}

\begin{figure}[tbp]
\centering
\includegraphics[height=0.55\textheight]{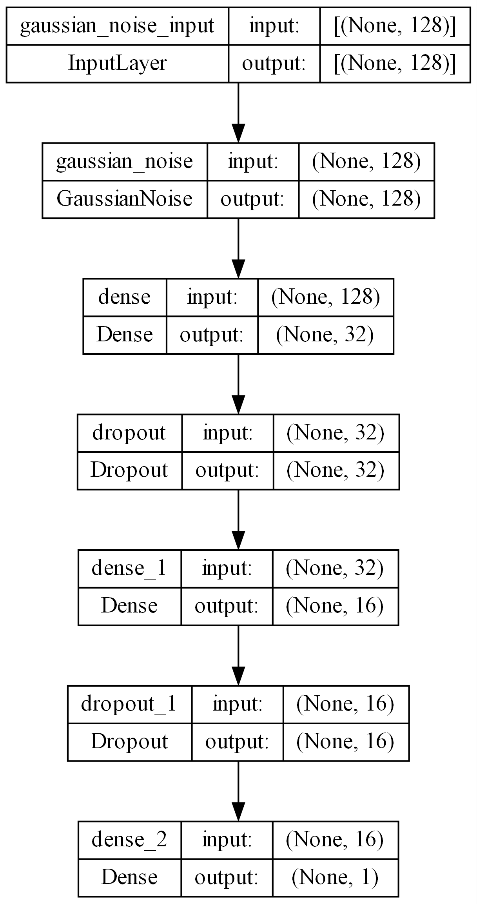}
\caption{\textbf{Neural classifier architecture.}}
\label{fig:classifier}
\end{figure}

\begin{table}[tbp]
\centering
\caption{\textbf{Network training parameters.}}
\label{tab:network_params}
\includegraphics[width=0.65\textwidth]{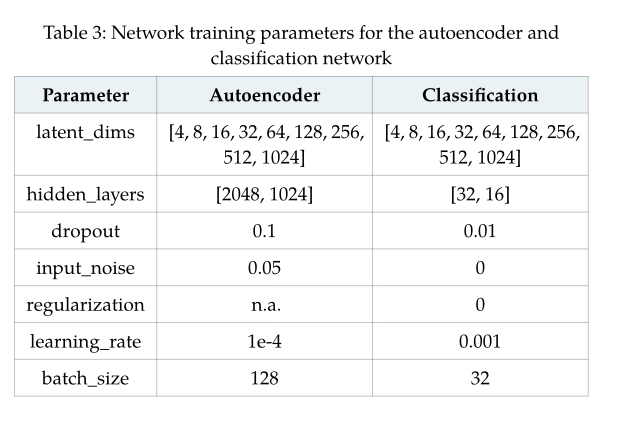}
\end{table}

\clearpage
\bibliographystyle{unsrtnat}
\bibliography{references}

\end{document}